\documentclass[sigconf,nonacm]{acmart}
\usepackage{graphicx,caption,subcaption}
\usepackage{amsmath}
\usepackage{amsfonts}
\usepackage{bbm}
\def\BibTeX{{\rm B\kern-.05em{\sc i\kern-.025em b}\kern-.08emT\kern-.1667em\lower.7ex\hbox{E}\kern-.125emX}}
    
\copyrightyear{2019}
\acmYear{2019}

\begin{document}

%
\title{On Expected Accuracy}

%
\author{Ozan \.Irsoy}
\email{oirsoy@bloomberg.net}
\affiliation{%
  \institution{Bloomberg L.P.}
  \streetaddress{731 Lexington Ave}
  \city{New York}
  \state{NY}
  \postcode{10022}
}

%
\renewcommand{\shortauthors}{Ozan \.Irsoy}

%
\begin{abstract}
We empirically investigate the (negative) expected accuracy as an alternative loss function to cross entropy (negative log likelihood) for classification tasks. Coupled with softmax activation, it has small derivatives over most of its domain, and is therefore hard to optimize. A modified, leaky version is evaluated on a variety of classification tasks, including digit recognition, image classification, sequence tagging and tree tagging, using a variety of neural architectures such as logistic regression, multilayer perceptron, CNN, LSTM and Tree-LSTM. We show that it yields comparable or better accuracy compared to cross entropy. Furthermore, the proposed objective is shown to be more robust to label noise.
\end{abstract}

%
\maketitle

\section{Introduction}

Classification is perhaps the most prominent supervised learning task in machine learning~\cite{alpaydin2009introduction}.
In classification, we are interested in assigning a given instance to a set of predetermined categories, based on
prior observations in our \emph{training} data. Typically, in classification, we use the maximum likelihood approach
to estimate model parameters~\cite{vapnik2013nature,millar2011maximum}. In this approach, we aim to find
the most likely model parameters that
could explain the observations in our training set. This leads to the popular negative log likelihood objective function.
However, there is an established mismatch in preeminent approaches:
Even though we optimize for the negative log likelihood, we still compare models on their (test) accuracy, or error
rate~\cite{kotsiantis2007supervised,weiss1990empirical,krizhevsky2009learning,nair2010rectified}. This leads us to ask:
why not optimize for accuracy directly? A simple answer would be that it is not differentiable, since it is not even continuous
at the decision boundary. Another reason might be the desirable properties of the likelihood approach: if the true class label
is probabilistic given by a joint distribution of instances and labels, the likelihood objective would converge to the actual
distribution, given enough data~\cite{kiefer1956consistency,banker1993maximum}. Still, in most settings we might actually only care about accuracy and think of log likelihood as
a surrogate function to it~\cite{witten2016data}. A mistake might have the same cost regardless of how close it is to the decision
boundary.

This is certainly not a new question. Prior work has investigated the notion of a surrogate loss function that upper bounds the
0-1 loss, with the assumption that, optimizing the surrogate risk results in a better true risk~\cite{pires2016multiclass, bartlett2006convexity}.
Alternatively, margin based loss functions such as the hinge loss in support vector machines provide alternatives to the probabilistic
log likelihood approaches~\cite{wu2007robust,wahba1999support, lin2004note}. Other work investigates the Fischer-consistent
loss functions (proper scoring rules), such as squared error loss or boosting loss~\cite{buja2005loss}.

In this work we investigate a very simplistic loss function: negative expected accuracy (or error rate). 
We show that even though we define the expectation over
the model distribution rather than the data distribution, this still gives us a loss function that
is close to the actual accuracy (or 0-1 loss). We subsequently see that this particular loss function
introduces difficulty in its optimization, and therefore further explore a leaky version of it. In a variety of experiments
that cover a wide range of architectures and settings, we compare it to the traditional log likelihood loss and examine
its strengths.

In Section~\ref{sec:method}, we provide the rigorous formulation of the expected accuracy and the leaky expected accuracy.
In Section~\ref{sec:prelim}, we perform preliminary experiments that compare different loss functions. Based on these preliminary results,
in Section~\ref{sec:setting} we lay out the further experimental setting and in Section~\ref{sec:main} we present the main
experimental results. Finally, we present our conclusions in Section~\ref{sec:conclusion}.

\section{Methodology}
\label{sec:method}

\begin{figure*}[th!]
    \centering
    \begin{subfigure}[t]{0.33\textwidth}
        \includegraphics[width=\textwidth]{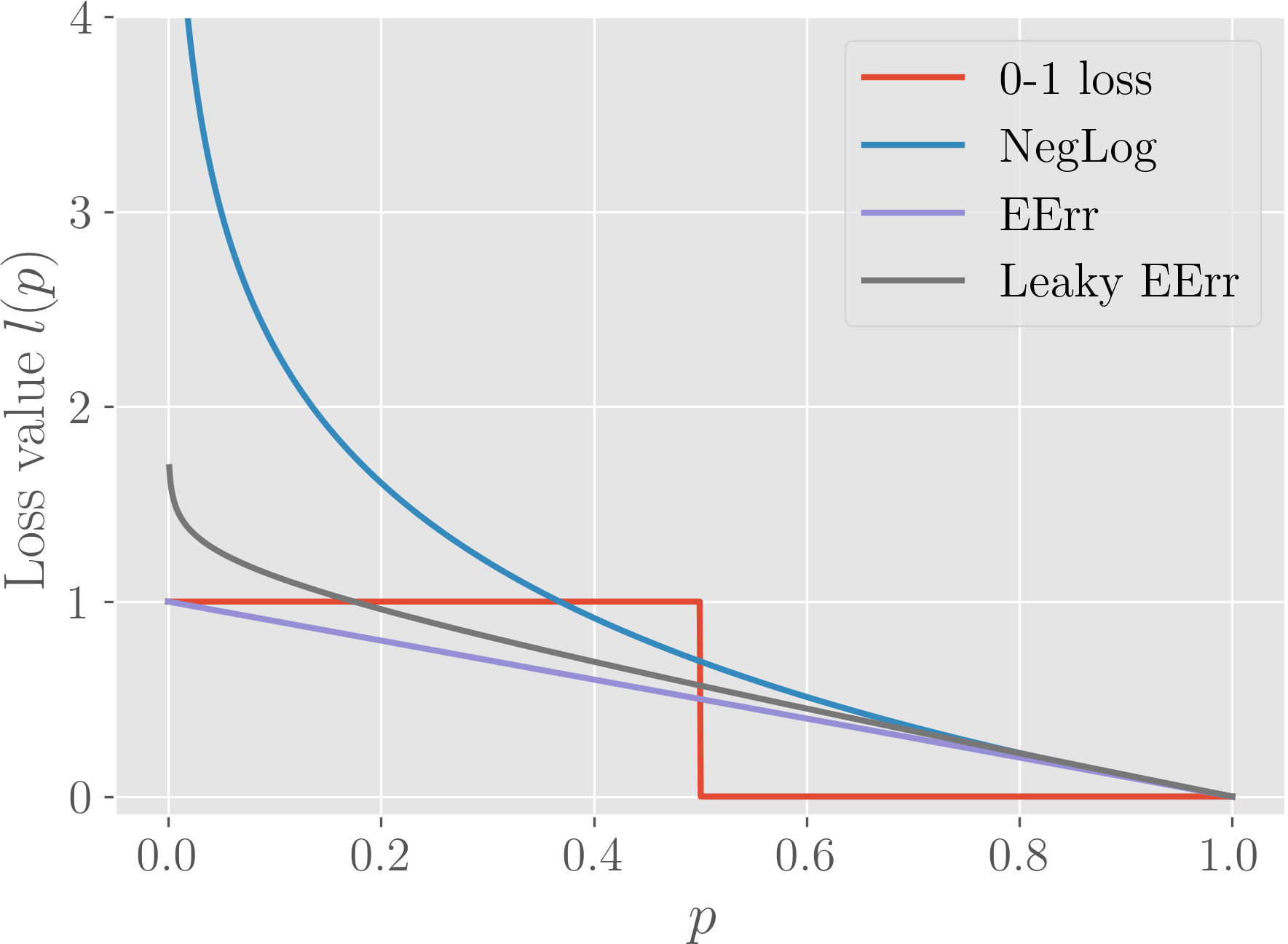}%
        \caption{Losses as functions of the probability of the true class.}
    \end{subfigure}%
    \begin{subfigure}[t]{0.33\textwidth}
        \includegraphics[width=\textwidth]{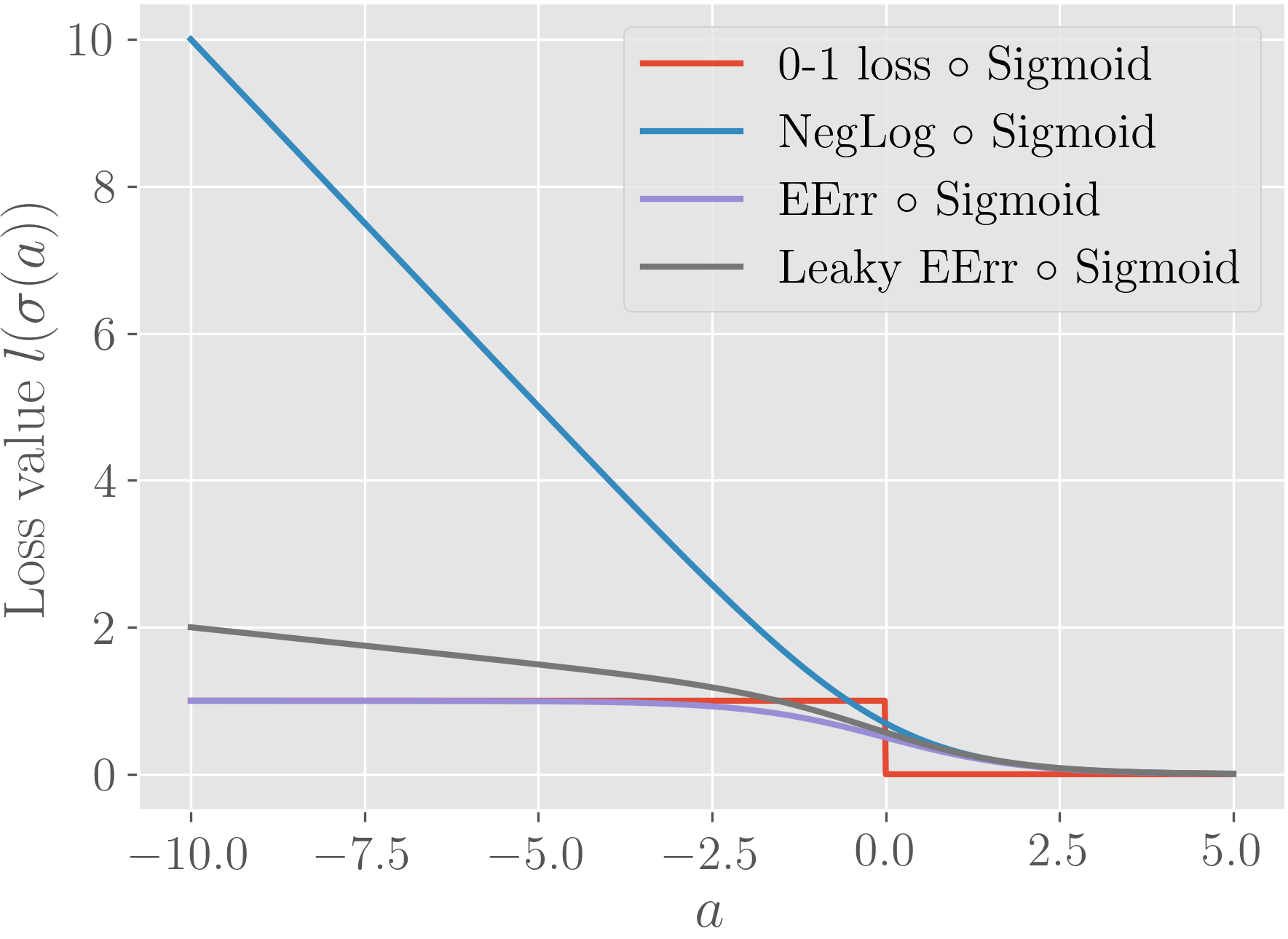}%
        \caption{Losses composed with sigmoid as functions of the pre-activation scores of the true class.}
    \end{subfigure}%
    \begin{subfigure}[t]{0.33\textwidth}
        \includegraphics[width=\textwidth]{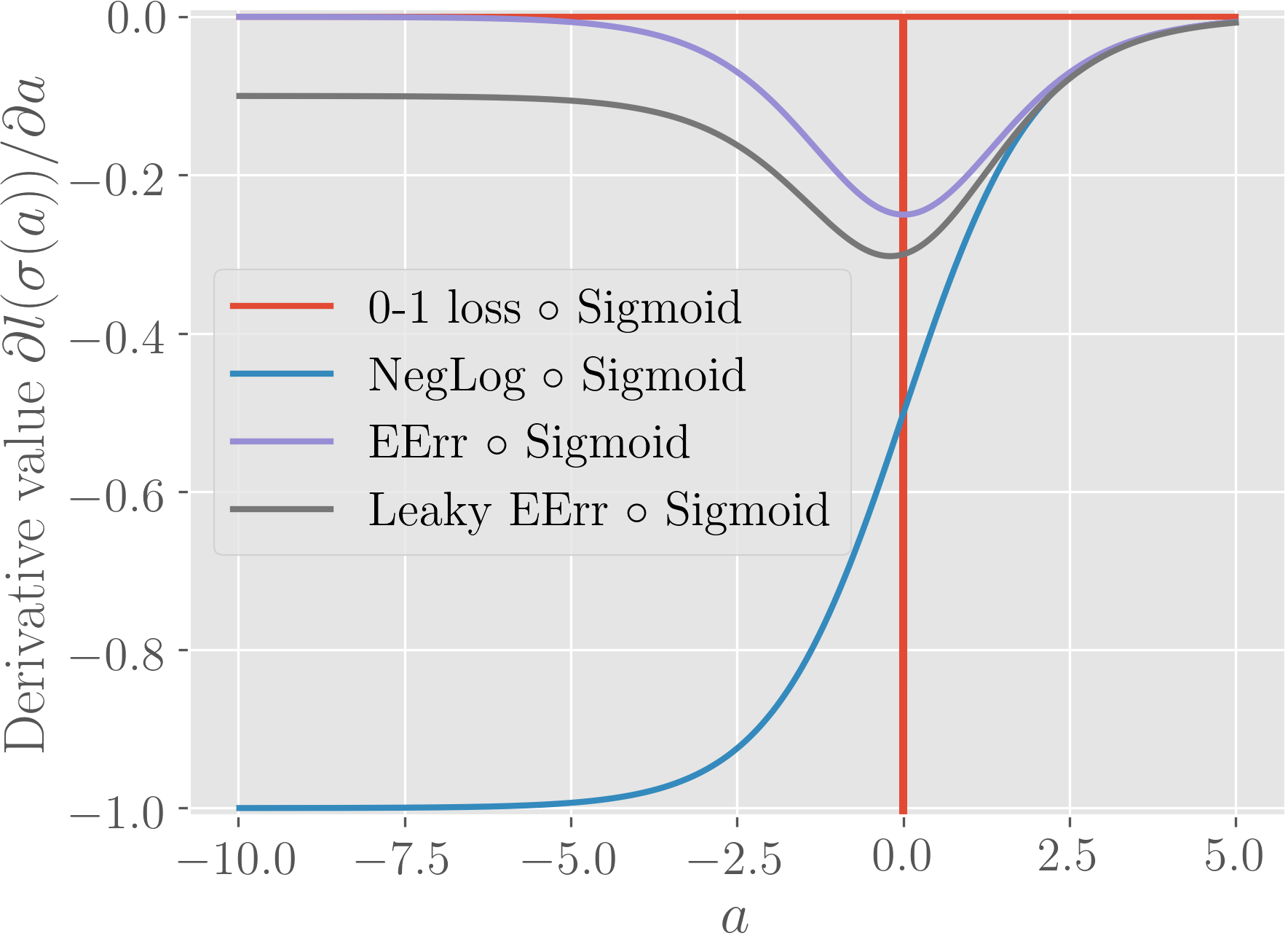}%
        \caption{Derivatives of loss-sigmoid compositions.}
    \end{subfigure}%
    \caption{Loss functions assuming a binary classification setting.}
    \label{fig:loss}
\end{figure*}

Consider a classification setting where a prediction function $f$ assigns a categorical distribution $\mathcal{Y}$,
to an input instance $x \in X$, and the final class assignment is done randomly 
by sampling a class label from $\mathcal{Y}$:
\begin{align}
    \mathcal{Y} &= f(x) &\mathcal{Y} \in [0,1]^k, |\mathcal{Y}| = 1\\
    y &\sim \mathcal{Y} & y \in \{1, \mathellipsis, k\}
\end{align}

This is slightly different than the traditional setting where the most likely class is picked ($y = \text{argmax}(\mathcal{Y})$).
However since exact accuracy would be discontinuous, the stochastic setting allows us to define a continuous and differentiable
proxy.

Given a dataset of instance - true label pairs $\{(x^{(i)}, r^{(i)})\}_{i}$, the \emph{expected accuracy} of the prediction
function $f$ would be:
\begin{align}
    \mathbb{E}[Acc] &= \dfrac{1}{N}\sum_i \mathbb{E}[\mathbbm{1}(y^{(i)} = r^{(i)})]\\
                    &= \dfrac{1}{N}\sum_i \mathbb{P}(y^{(i)} = r^{(i)}) \\
                    &= \dfrac{1}{N}\sum_i \mathcal{Y}^{(i)}_{r^{(i)}}
\end{align}
This is simply the sum of all probabilities assigned to the correct class labels (up to a constant factor of  $1/N$).

We can negate this quantity to turn into a loss function (-$\frac{1}{N}\sum_i \mathcal{Y}^{(i)}_{r^{(i)}}$).
We can additionally translate it to get the \emph{expected error rate} ((1-$\frac{1}{N}\sum_i \mathcal{Y}^{(i)}_{r^{(i)}}$)),
which would yield the same objective up to a constant additive factor.

\subsection{Comparison to negative log likelihood}

Negative expected accuracy as defined above looks similar to negative log likelihood except that we sum the probabilities
themselves rather than their logs. Both loss functions optimize for high probability values assigned to the correct class,
but weighted differently.

\textbf{Surrogate for the 0-1 loss.} If we look at the task of classification from an optimization point of view, we can describe
the approach as follows:
\begin{enumerate}
    \item Our main goal is to optimize for the test set accuracy.
    \item Since we cannot optimize over unseen data, we settle for optimizing for the training set accuracy and hope to have good
    test set accuracy as a side effect.
    \item Since we cannot optimize for accuracy using a gradient-based method (due to its nondifferentiability), we settle for
    optimizing a differentiable surrogate function that approximates it well 
    enough.\footnote{There is an argument that a better training objective surrogate (or even the exact accuracy)
    could be worse for test accuracy. We discuss this in the final section.}
\end{enumerate}

In this regard, we can compare both losses with respect to the 0-1 loss (error rate for a single instance) as a function of the
probability value assigned to the true class label. We visualize the functions in Figure~\ref{fig:loss} (a). Negative log likelihood
diverges from 0-1 loss as we approach 0. It values an increase in probability values (of the true class),
say, from 0.1 to 0.2 more than an increase from 0.45 to 0.55,
whereas both would be similar for the expected accuracy. We posit that instead of prioritizing correction of those instances
that we perform very poorly on, by weighing probability errors equally, we might just be able to push more instances to the other side
of the decision boundary. Note that in the cases that we can perform well on the training set for the negative log likelihood, both
functions behave similarly.

\textbf{As functions of pre-activations.} Commonly the softmax activation (or sigmoid in the binary classification case) is used to convert
unbounded scores (pre-activations) to probability values. We can consider the composition of loss functions and the softmax as a function
of these pre-activations $a$, which gives us another view. We visualize the compositions as such in Figure~\ref{fig:loss} (b).

Logarithm and the exponential within sigmoid cancel each other asymptotically for -log(sigmoid($\cdot$)) for negative values of $a$.
This approximately linear behavior allows it to have (absolutely) large derivatives 
($\approx -1$) which is desirable for its optimization. On the other hand, expected error rate, coupled with the sigmoid has an
asymptotically zero derivative around the negative region, which potentially make it hard to optimize. For the instances that we are the
most incorrect on, progress could be very little. Still, the main motivating
idea behind it is to provide more incentive (larger absolute derivatives) over the instances that are closer to the decision boundary,
to grab the lower hanging fruit first.

As we will see in the later sections, difficulty of optimizing the negative expected accuracy will indeed present itself as a practical
issue. To combat this, we explore a leaky version of it, by combining it with the traditional log likelihood function:
\begin{align}
   L &= -\dfrac{1}{N}\sum_i (\mathcal{Y}^{(i)}_{r^{(i)}} + \alpha \log \mathcal{Y}^{(i)}_{r^{(i)}})
   \label{eq:leaky}
\end{align}
for some small value of $\alpha$. We use $\alpha=0.1$ in this work. As seen in Figure~\ref{fig:loss}, this gives us a similar
curve while having a nonzero asymptotic derivative in the negative region.

\textbf{Bayes optimal predictors.} In general, in classification we assume the true label of an instance is a random variable rather
than a deterministic value, since $(x, r)$ is assumed to be from a joint distribution.
Negative log likelihood objective has the desirable property that the predicted conditional distribution
$(\mathcal{Y}|\mathcal{X})$ converges to the true distribution $(\mathcal{R}|\mathcal{X})$ as we have more observations. This does not 
hold for the expected accuracy. In fact the Bayes optimal predictor for it is to assign a one-hot probability distribution which marks
the most likely class. Maximum likelihood approach strives for matching the predicted and true \emph{distributions} of labels,
where expected accuracy wants to simply improve the counts for \emph{matching class predictions}. Since from an accuracy perspective
the best label that we can predict is the one that has the most instances in the population, such a Bayes optimal predictor is intuitive
for its objective.

\textbf{Noisy labels.} Since negative likelihood diverges the most from 0-1 loss in the most negative region, we hypothesize that
the impact of the proposed alternative will be the most apparent in the noisy label setting where each instance has a probability of its label being flipped.
This setting simulates practical issues such as annotation errors.

\section{Experiments}
\subsection{Preliminary Experiments}
\label{sec:prelim}

For preliminary experimentation, we use the logistic regression method over the MNIST digit recognition dataset, which
has 60k train and 10k test instances of digits which are of 28$\times$28 dimensional. The task is to classify each digit
into one of the ten classes. We train the model over 200 epochs using the Adam update rule~\cite{kingma2014adam} with a learning
rate of 1e-4.

\begin{figure}[t!]
    \centering
    \includegraphics[width=\columnwidth]{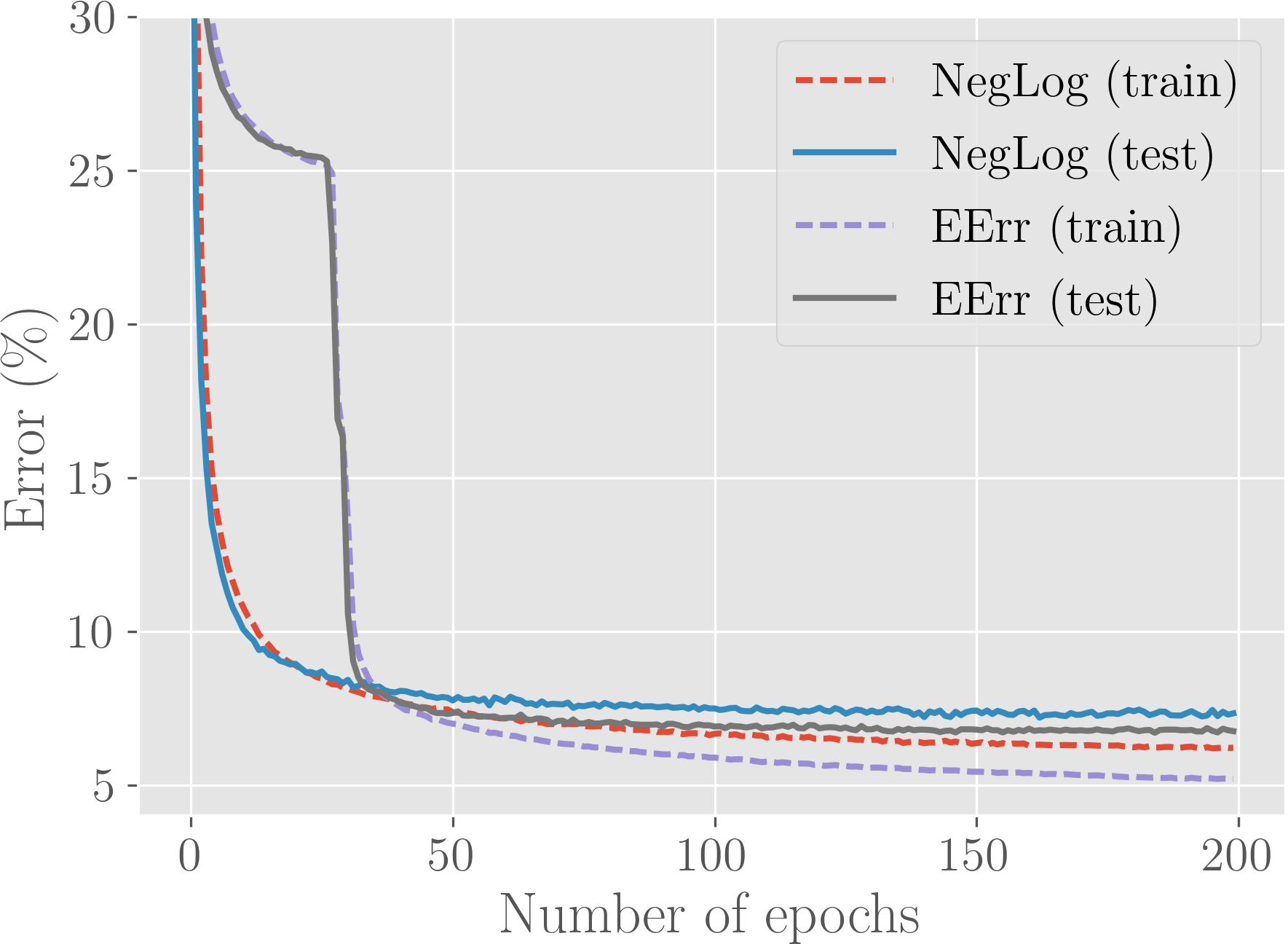}
    \caption{Logistic regression on MNIST.}
    \label{fig:logreg-mnist}
\end{figure}

Results are given in Figure~\ref{fig:logreg-mnist} as training and test curves. We see that in terms of both training and test
performance, expected accuracy performs better. However, as we suspected, we observe a temporary plateauing of performance in
the early stages of training for the expected accuracy.

To account for randomness, and investigate consistency of the behavioral patterns,
we perform 10-fold cross-validation by splitting the entire training set into training-development
partitions of ratio 9:1. Development set is used for early stopping for a patience value of 15 epochs.

\begin{figure}[b!]
    \centering
    \includegraphics[width=\columnwidth]{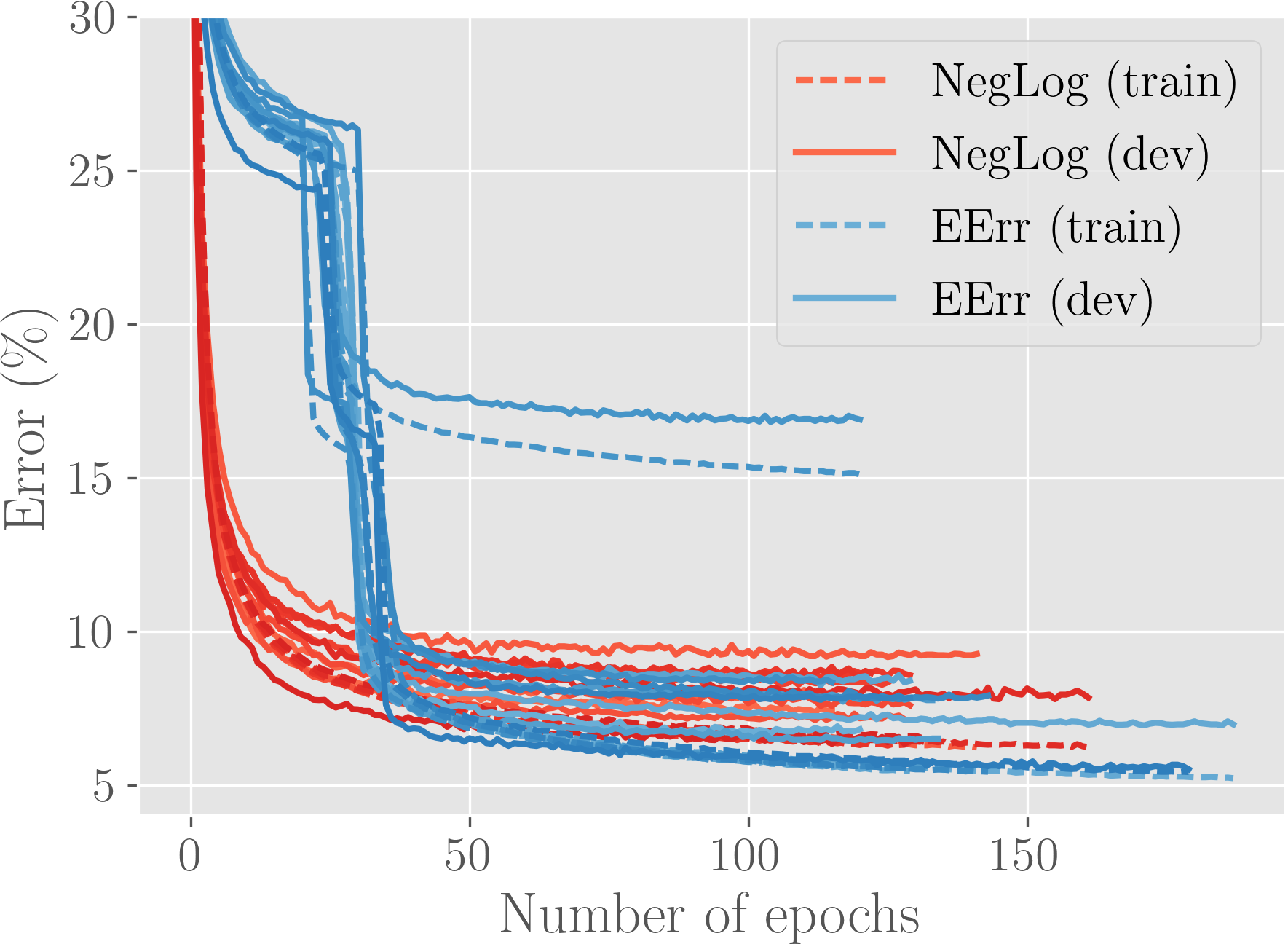}
    \caption{Logistic regression on MNIST over ten folds.}
    \label{fig:logreg-mnist-repl}
\end{figure}

Results for the replicated experiments are given in Figure~\ref{fig:logreg-mnist-repl}. We observe that initial plateauing
is consistent across different runs. Furthermore, even though we see good performance compared to negative log likelihood
for many runs, there is a particular run that the expected accuracy cannot overcome the initial plateau, resulting in
a suboptimal performance by a wide margin.

\begin{figure}[t!]
    \centering
    \includegraphics[width=\columnwidth]{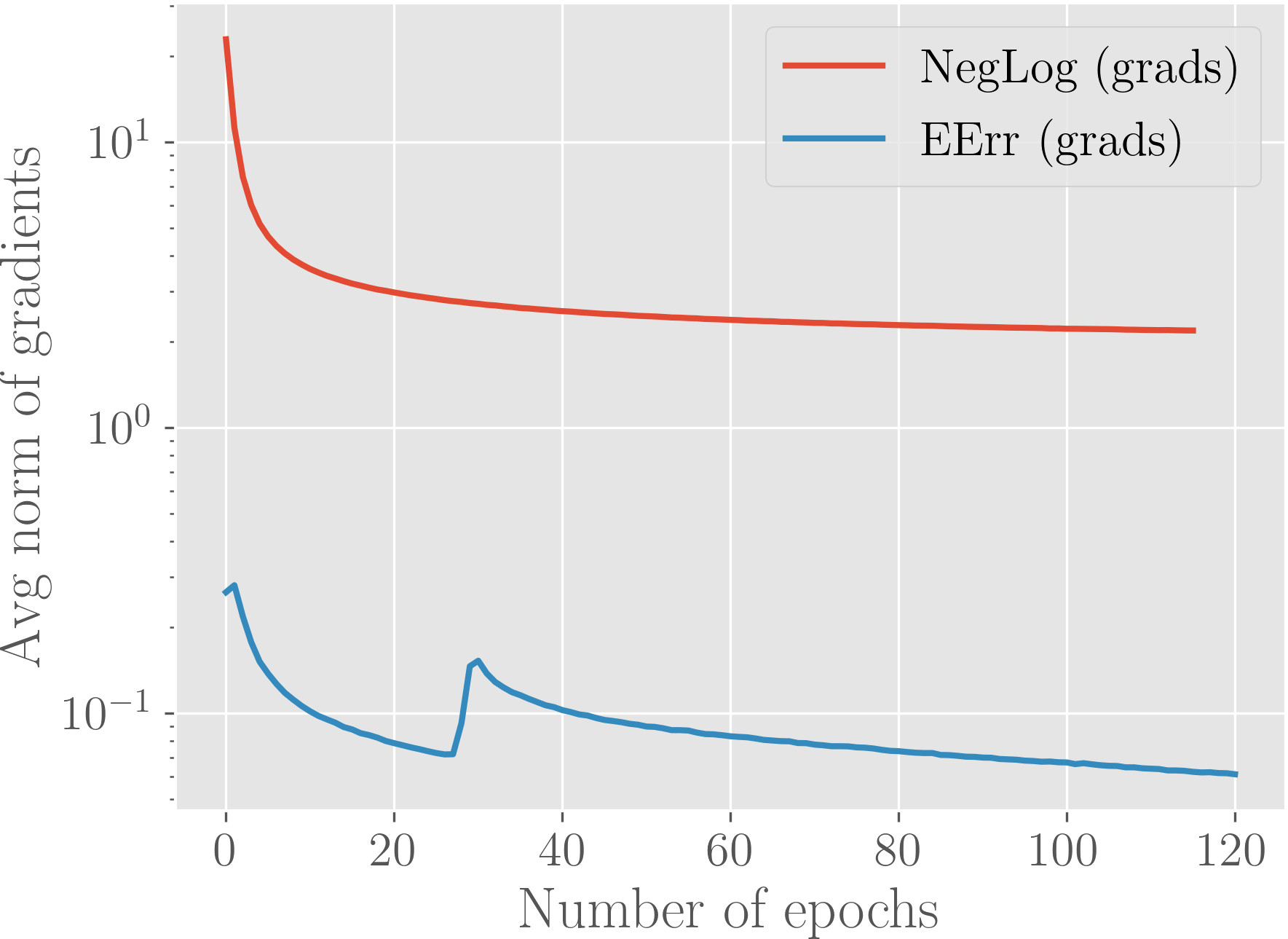}
    \caption{Average norms of gradients w.r.t pre-activations.}
    \label{fig:mnist-grads}
\end{figure}

\textbf{Magnitudes of gradients.} A possible culprit that would explain the early plateauing is simply the low magnitudes
of gradients for the loss  function of interest. We plot the average norms of the gradients of both losses (composed with
the softmax) with respect to the softmax pre-activations in Figure~\ref{fig:mnist-grads} for one of the runs.

As we confirm a two order of magnitude difference in the norms, we rerun the experiments with a learning rate
of 1e-2 for the negative expected accuracy loss, since merely having
a larger learning rate might resolve getting stuck early on. However we observe a behavior that is very similar to
Figure~\ref{fig:logreg-mnist-repl} (henceforth, the plot is ommitted), which confirms that the issue is more fundamental.
This justifes  the  use of the leaky expected accuracy as defined in Equation~\ref{eq:leaky} as a simple workaround.

\subsection{Main Experimental Setting}
\label{sec:setting}

\textbf{Architectures.} For further experiments we explore several classification related tasks using various architectures:
\begin{itemize}
    \item \textbf{Multilayer perceptron (MLP).} We use a three hidden layer feedforward neural network with ReLU
    activations~\cite{nair2010rectified}. Number of hidden units are set to 300, 200 and 100 for each layer respectively.
    We use dropout regularization for both the input as well as the hidden layers in which we randomly
    drop each unit with probability $p$~\cite{JMLR:v15:srivastava14a}.
    \item \textbf{Convolutional neural network (CNN).} As a CNN we use the ResNet18 deep residual network
    architecture~\cite{he2016deep}. To combat overfitting, we provide small random translations or horizontal flips of
    each training image to the network.
    \item \textbf{Bidirectional LSTM.} For sequence tagging tasks, we use a bidirectional long short-term memory
    architecture~\cite{hochreiter1997long, schuster1997bidirectional}. We use 100 hidden units / memory cells in each
    direction.
    \item \textbf{Tree LSTM.} For tree tagging (e.g. sentiment classification over parse trees of sentences) we use a
    tree LSTM architecture~\cite{tai2015improved} which generalizes the traditional LSTM such that it can operate over tree structures.
    Again, we fix the number of hidden units to 100.
\end{itemize}

For architectures that operate over textual data, we represent each word using a dense word embedding. To this end,
we use the pretrained 300-dimensional Glove word embeddings~\cite{pennington2014glove}.

\textbf{Data.} For logistic regression, we use six relatively small datasets from the UCI repository~\cite{Dua:2017}:
\emph{magic}~\cite{bock2004methods}, \emph{musk2}~\cite{dietterich1994comparison}, \emph{pima}~\cite{smith1988using},
\emph{polyadenylation}, \emph{ringnorm}~\cite{breiman1996bias}, \emph{satellite47}.
Number of instances, dimensionality and number of class labels for each dataset are shown in Table~\ref{tab:uci}.

For the MLP architecture, we use the MNIST dataset which we used for our preliminary experiments.

For the CNN architecture (ResNet) we use the CIFAR10 dataset which poses an image classification task, where
each 32$\times$32$\times$3 image is to be classified into one of the ten categories~\cite{krizhevsky2009learning}.

For sequence tagging tasks (in which we use the LSTM architecture), we focus on part-of-speech tagging (POS) and
named entity recognition (NER)~\cite{huang2015bidirectional}. We use Penn Treebank (PTB) dataset 
\cite{marcus1993building} for POS and CoNLL 2003
dataset \cite{tjong2003introduction} for NER.

For tree tagging, we use the Stanford Sentiment Treebank (SSTB) \cite{socher2013recursive}.
SSTB includes a supervised sentiment label for every node in the binary parse tree of each sentence.
Therefore, not only the sentences but every possible phrase within a sentence is labeled with a
sentiment score.

See Table~\ref{tab:other} for a breakdown of the datasets.

For a subset of the data, we also experiment in the noisy label setting where we randomly assign random labels
(with a probability of 0.05) to each instance in the training and development sets. Noisy versions  of the
dataset are denoted with an asterisk in the results.

\textbf{Learning.} For each task, we use the Adam update rule~\cite{kingma2014adam} and Xavier random initialization
of parameters~\cite{glorot2010understanding}. Since batching is nontrivial, for Tree LSTM (over SSTB), we use the
purely online setting of stochastic gradient descent (SGD), whereas for every other task we use minibatched training.
For all tasks we perform early stopping, i.e. we pick the best iteration out of all epochs based on the development
set performance. Additionally, we tune the learning rate (and dropout rate when appropriate)
over the same development set. For logistic regression, we use a minimum number of 100 epochs after which we start applying an
15 epoch patience rule (a lack of improvement for 15 epoch over the development set ends the run). This is 
because logistic regression is the least costly method and the datasets are small. For MLP and LSTMs, we apply
a 30 epoch patience with no minimum or maximum number of epochs. For CNNs and Tree LSTMs, we apply 200 epochs
without a patience value. These hyperparameters are intentionally left different to cover a wider range of settings.

\textbf{Replication.} For purposes of replication and to account for extraneous randomness such as data splits,
initialization, or the order of instances in SGD, we perform cross-validation (CV). After the original test set is
left apart, we randomly split the remaining data into training and development partitions. For MNIST and CIFAR,
we use 10-fold CV (there is no development partition readily available). For PTb and SSTB we first combine
the original training and development partitions into a bigger set and then apply 5-fold CV, by respecting
the original training and development set sizes of each dataset. For UCI datasets, we apply 5$\times$2-CV, 
which simply reapplies 2-fold CV five times. The only exception to cross-validation is the CoNLL03 data:
Since training / development / test partitions are temporally ordered, shuffling and resplitting is not possible.
For this data we replicate 5 different random initializations over the same partition.

Finally, we report average accuracy / error rates (with standard deviations) over the test set and compare them
using the paired t-test.

\begin{table}[tb!]
    \centering
    \begin{tabular}{l|rrc}
         Data & Instances & Dims & Classes  \\
         \hline
         magic & 19020 & 10 & 2 \\
         musk2 & 6598 & 166 & 2\\
         pima & 768 & 8 & 2\\
         polyadenylation & 6371 & 169 & 2 \\
         ringnorm & 7400 & 20 & 2 \\
         satellite47 & 2134 & 36 & 2\\
    \end{tabular}
    \caption{UCI Datasets}
    \label{tab:uci}
\end{table}

\begin{table}[tb!]
    \centering
    \begin{tabular}{l|rcc}
         Data & \begin{tabular}{@{}c@{}}Instances / \\ Sentences\end{tabular} & Dims & Classes  \\
         \hline
         MNIST & 70000 & 28 $\times$ 28 & 10 \\
         CIFAR10 & 60000 & 32 $\times$ 32 $\times$ 3 & 10\\
         PTB & 49208 & - & 45\\
         CoNLL03 & 22137 & - & 9 \\
         SSTB & 11855 & - & 5 \\
    \end{tabular}
    \caption{Other Datasets}
    \label{tab:other}
\end{table}

\subsection{Main Experiments}
\label{sec:main}

\begin{table}[tb!]
    \centering
    \begin{tabular}{l|rrr}
         Data & NegLog & EErr & LEErr\\
         \hline
         magic & 20.84 $\pm$ 0.26 & \underline{20.54} $\pm$ 0.27 & \underline{20.52} $\pm$ 0.18\\
         musk2 & 5.84 $\pm$ 0.64 & \underline{5.41} $\pm$ 0.36 & \underline{5.21} $\pm$ 0.53\\
         pima & \underline{23.54} $\pm$ 1.41 & 25.06 $\pm$ 2.14 & \underline{23.50} $\pm$ 0.91\\
         polya & \underline{22.66} $\pm$ 0.38 & 23.35 $\pm$ 0.55 & \underline{22.82} $\pm$ 0.30\\
         ringn & 23.48 $\pm$ 0.23 & \underline{22.67} $\pm$ 0.41 & \underline{22.78} $\pm$ 0.28\\
         sat47 & \underline{16.74} $\pm$ 0.80 & \underline{16.21} $\pm$ 0.56 & \underline{16.49} $\pm$ 0.88\\
    \end{tabular}
    \caption{Error rates of different loss functions with logistic regression (with standard deviations).}
    \label{tab:logreg}
\end{table}

Logistic regression experiments using the UCI datasets are shown in Table~\ref{tab:logreg}. We report mean
error rate with the standard deviation across replications. Best results, as well as the ones that perform
no worse than the best results in a statistically significant fashion are shown underlined ($\alpha=0.05$).
Comparing leaky expected error (LEErr) and negative log likelihood (NegLog), we see three wins for
LEErr (\emph{magic, musk2, ringnorm}) and three ties (\emph{pima, polyadenylation, satellite47}).
We see that  LEErr consistently performs best (or no worse that best). Unmodified expected error rate (EErr)
has four instances in which it is no worse than the best, however for two datasets, it is significantly
worse than the other two (\emph{pima, polyadenylation}). For later experiments we omit EErr.

\begin{table}[tb!]
    \centering
    \begin{tabular}{l|rrr}
         Data & NegLog & LEErr\\
         \hline
         MNIST  & 1.49 $\pm$ 0.08 & \underline{1.40} $\pm$ 0.08\\
         MNIST* & 1.77 $\pm$ 0.08 & \underline{1.61} $\pm$ 0.08\\
    \end{tabular}
    \caption{Error rates of MLP using different loss functions.}
    \label{tab:mlp}
\end{table}

Table~\ref{tab:mlp} lists the results for the MLP architecture using MNIST. We observe that LEErr outperforms
NegLog by a slight but (statistically) significant margin. When we rerun the experiment using the noisy label case (denoted
as MNIST*), we observe a similar result with an increase in the margin. This is in line with our hypothesis about the
differences of the two function possibly being more noticeable in the noisy labeling setting since more instances will lie
in the negative region.

\begin{table}[tb!]
    \centering
    \begin{tabular}{l|rrr}
         Data & NegLog & LEErr\\
         \hline
         CIFAR10  & \underline{92.20} $\pm$ 0.25 & \underline{92.39} $\pm$ 0.20\\
    \end{tabular}
    \caption{Accuracy of ResNet18 using different loss functions.}
    \label{tab:cnn}
\end{table}

ResNet18 results over CIFAR10 are presented in Table~\ref{tab:cnn}. For this setting, we observe no discernible
difference with the two approaches. On average, LEErr performs about single standard deviation better than
NegLog, however the difference is not statistically significant.

\begin{table}[tb!]
    \centering
    \begin{tabular}{l|rrr}
         Data & NegLog & LEErr\\
         \hline
         PTB      & 96.82 $\pm$ 0.05 & \underline{96.95} $\pm$ 0.03\\
         CoNLL03  & \underline{97.36} $\pm$ 0.06 & \underline{97.37} $\pm$ 0.07\\
         CoNLL03* & 97.12 $\pm$ 0.12 & \underline{97.37} $\pm$ 0.06\\
    \end{tabular}
    \caption{Accuracies of bidirectional LSTMs.}
    \label{tab:lstm}
\end{table}

Results for sequence tagging using bidirectional LSTMs are given in Table~\ref{tab:lstm}.
For part-of-speech tagging over PTB, we see a statistically significant improvement using LErr
(96.95 vs 96.82). For named entity recognition over CoNLL03 however we see a very close tie.
When we inject noise to the labels, there is no degradation  for LErr (the test accuracy stays
at 97.37) whereas NegLog drops from 97.36 to 97.12. In that setting, the difference between
NegLog and LEErr is significant.

\begin{table}[tb!]
    \centering
    \begin{tabular}{l|rrr}
         Data & NegLog & LEErr\\
         \hline
         SSTB (sent)  & \underline{48.94} $\pm$ 0.82 & \underline{48.55} $\pm$ 0.80\\
         SSTB (phrase)  & \underline{82.11} $\pm$ 0.09 & \underline{82.03} $\pm$ 0.13\\
    \end{tabular}
    \caption{Accuracy of TreeLSTM using different loss functions.}
    \label{tab:tlstm}
\end{table}

Finally, results for sentiment classification over binary parse trees are demonstrated in Table~\ref{tab:tlstm}.
Since the data contains sentiment labels all phrases (all tree nodes) as well as sentences (only root nodes),
we can evaluate the accuracy for both. For both measurements we see a tie: LEErr performs slightly worse than
NegLog, however the difference is not significant.

\section{Conclusion and Future Work}
\label{sec:conclusion}

We experimentally investigate the expected accuracy / error rate (and in particular, its leaky version)
as an alternative classification objective over a number of architectures and tasks. For some settings, we observe
improvements over log likelihood, such as logistic regression, multilayer perceptron and sequence tagging with RNNs.
For others, it performs comparably, e.g. for CNNs and Tree LSTMs. We  find the results promising since LEErr overall
performs better or no  worse than NegLog.

One of the main motivations behind expected accuracy is to provide a more faithful approximation to accuracy.
However there is a chance that optimizing for training accuracy (or expected accuracy) to be worse
for generalization accuracy. For instance, this is an argument for the margin based loss approaches,
since having a large margin around the decision boundary tends to improve generalization (even though accuracy
itself does not have any margin) \cite{lin2004note}. In this work, we compare loss functions over the test set
to ensure that their generalization performance is evaluated.

Log likelihood is perhaps the most commonly used probabilistic objective for classification, and is well studied.
Many of the recent innovations in deep learning that led to being able to train better models, such as
improved regularization~\cite{JMLR:v15:srivastava14a}, better activation functions~\cite{nair2010rectified}, or
improved update rules~\cite{kingma2014adam} are studied using the cross entropy
classification objective, therefore expected to synergize well with it. Similarly, recent results on loss surfaces
or convergence dynamics of neural networks use softmax with log likelihood losses \cite{NIPS2018_7875, NIPS2017_6662}.
We believe that future work that uses the (leaky) expected accuracy objective could discover more compatible
hyperparameters with improved performance.

%
\bibliographystyle{ACM-Reference-Format}
\bibliography{ref}


\begin{thebibliography}{35}


\ifx \showCODEN    \undefined \def \showCODEN     #1{\unskip}     \fi
\ifx \showDOI      \undefined \def \showDOI       #1{#1}\fi
\ifx \showISBNx    \undefined \def \showISBNx     #1{\unskip}     \fi
\ifx \showISBNxiii \undefined \def \showISBNxiii  #1{\unskip}     \fi
\ifx \showISSN     \undefined \def \showISSN      #1{\unskip}     \fi
\ifx \showLCCN     \undefined \def \showLCCN      #1{\unskip}     \fi
\ifx \shownote     \undefined \def \shownote      #1{#1}          \fi
\ifx \showarticletitle \undefined \def \showarticletitle #1{#1}   \fi
\ifx \showURL      \undefined \def \showURL       {\relax}        \fi
\providecommand\bibfield[2]{#2}
\providecommand\bibinfo[2]{#2}
\providecommand\natexlab[1]{#1}
\providecommand\showeprint[2][]{arXiv:#2}

\bibitem[\protect\citeauthoryear{Alpaydin}{Alpaydin}{2009}]%
        {alpaydin2009introduction}
\bibfield{author}{\bibinfo{person}{Ethem Alpaydin}.}
  \bibinfo{year}{2009}\natexlab{}.
\newblock \bibinfo{booktitle}{\emph{Introduction to machine learning}}.
\newblock \bibinfo{publisher}{MIT press}.
\newblock


\bibitem[\protect\citeauthoryear{Banker}{Banker}{1993}]%
        {banker1993maximum}
\bibfield{author}{\bibinfo{person}{Rajiv~D Banker}.}
  \bibinfo{year}{1993}\natexlab{}.
\newblock \showarticletitle{Maximum likelihood, consistency and data
  envelopment analysis: a statistical foundation}.
\newblock \bibinfo{journal}{\emph{Management science}} \bibinfo{volume}{39},
  \bibinfo{number}{10} (\bibinfo{year}{1993}), \bibinfo{pages}{1265--1273}.
\newblock


\bibitem[\protect\citeauthoryear{Bartlett, Jordan, and McAuliffe}{Bartlett
  et~al\mbox{.}}{2006}]%
        {bartlett2006convexity}
\bibfield{author}{\bibinfo{person}{Peter~L Bartlett},
  \bibinfo{person}{Michael~I Jordan}, {and} \bibinfo{person}{Jon~D McAuliffe}.}
  \bibinfo{year}{2006}\natexlab{}.
\newblock \showarticletitle{Convexity, classification, and risk bounds}.
\newblock \bibinfo{journal}{\emph{J. Amer. Statist. Assoc.}}
  \bibinfo{volume}{101}, \bibinfo{number}{473} (\bibinfo{year}{2006}),
  \bibinfo{pages}{138--156}.
\newblock


\bibitem[\protect\citeauthoryear{Bock, Chilingarian, Gaug, Hakl, Hengstebeck,
  Ji{\v{r}}ina, Klaschka, Kotr{\v{c}}, Savick{\`y}, Towers, et~al\mbox{.}}{Bock
  et~al\mbox{.}}{2004}]%
        {bock2004methods}
\bibfield{author}{\bibinfo{person}{RK Bock}, \bibinfo{person}{A Chilingarian},
  \bibinfo{person}{M Gaug}, \bibinfo{person}{F Hakl}, \bibinfo{person}{Th
  Hengstebeck}, \bibinfo{person}{M Ji{\v{r}}ina}, \bibinfo{person}{J Klaschka},
  \bibinfo{person}{E Kotr{\v{c}}}, \bibinfo{person}{P Savick{\`y}},
  \bibinfo{person}{S Towers}, {et~al\mbox{.}}} \bibinfo{year}{2004}\natexlab{}.
\newblock \showarticletitle{Methods for multidimensional event classification:
  a case study using images from a Cherenkov gamma-ray telescope}.
\newblock \bibinfo{journal}{\emph{Nuclear Instruments and Methods in Physics
  Research Section A: Accelerators, Spectrometers, Detectors and Associated
  Equipment}} \bibinfo{volume}{516}, \bibinfo{number}{2-3}
  (\bibinfo{year}{2004}), \bibinfo{pages}{511--528}.
\newblock


\bibitem[\protect\citeauthoryear{Breiman}{Breiman}{1996}]%
        {breiman1996bias}
\bibfield{author}{\bibinfo{person}{Leo Breiman}.}
  \bibinfo{year}{1996}\natexlab{}.
\newblock \showarticletitle{Bias, variance, and arcing classifiers}.
\newblock  (\bibinfo{year}{1996}).
\newblock


\bibitem[\protect\citeauthoryear{Buja, Stuetzle, and Shen}{Buja
  et~al\mbox{.}}{2005}]%
        {buja2005loss}
\bibfield{author}{\bibinfo{person}{Andreas Buja}, \bibinfo{person}{Werner
  Stuetzle}, {and} \bibinfo{person}{Yi Shen}.} \bibinfo{year}{2005}\natexlab{}.
\newblock \showarticletitle{Loss functions for binary class probability
  estimation and classification: Structure and applications}.
\newblock \bibinfo{journal}{\emph{Working draft, November}}
  \bibinfo{volume}{3} (\bibinfo{year}{2005}).
\newblock


\bibitem[\protect\citeauthoryear{Dheeru and Karra~Taniskidou}{Dheeru and
  Karra~Taniskidou}{2017}]%
        {Dua:2017}
\bibfield{author}{\bibinfo{person}{Dua Dheeru} {and} \bibinfo{person}{Efi
  Karra~Taniskidou}.} \bibinfo{year}{2017}\natexlab{}.
\newblock \bibinfo{title}{{UCI} Machine Learning Repository}.
\newblock
\newblock
\urldef\tempurl%
\url{http://archive.ics.uci.edu/ml}
\showURL{%
\tempurl}


\bibitem[\protect\citeauthoryear{Dietterich, Jain, Lathrop, and
  Lozano-Perez}{Dietterich et~al\mbox{.}}{1994}]%
        {dietterich1994comparison}
\bibfield{author}{\bibinfo{person}{Thomas~G Dietterich},
  \bibinfo{person}{Ajay~N Jain}, \bibinfo{person}{Richard~H Lathrop}, {and}
  \bibinfo{person}{Tomas Lozano-Perez}.} \bibinfo{year}{1994}\natexlab{}.
\newblock \showarticletitle{A comparison of dynamic reposing and tangent
  distance for drug activity prediction}. In \bibinfo{booktitle}{\emph{Advances
  in Neural Information Processing Systems}}. \bibinfo{pages}{216--223}.
\newblock


\bibitem[\protect\citeauthoryear{Glorot and Bengio}{Glorot and Bengio}{2010}]%
        {glorot2010understanding}
\bibfield{author}{\bibinfo{person}{Xavier Glorot} {and} \bibinfo{person}{Yoshua
  Bengio}.} \bibinfo{year}{2010}\natexlab{}.
\newblock \showarticletitle{Understanding the difficulty of training deep
  feedforward neural networks}. In \bibinfo{booktitle}{\emph{Proceedings of the
  thirteenth international conference on artificial intelligence and
  statistics}}. \bibinfo{pages}{249--256}.
\newblock


\bibitem[\protect\citeauthoryear{He, Zhang, Ren, and Sun}{He
  et~al\mbox{.}}{2016}]%
        {he2016deep}
\bibfield{author}{\bibinfo{person}{Kaiming He}, \bibinfo{person}{Xiangyu
  Zhang}, \bibinfo{person}{Shaoqing Ren}, {and} \bibinfo{person}{Jian Sun}.}
  \bibinfo{year}{2016}\natexlab{}.
\newblock \showarticletitle{Deep residual learning for image recognition}. In
  \bibinfo{booktitle}{\emph{Proceedings of the IEEE conference on computer
  vision and pattern recognition}}. \bibinfo{pages}{770--778}.
\newblock


\bibitem[\protect\citeauthoryear{Hochreiter and Schmidhuber}{Hochreiter and
  Schmidhuber}{1997}]%
        {hochreiter1997long}
\bibfield{author}{\bibinfo{person}{Sepp Hochreiter} {and}
  \bibinfo{person}{J{\"u}rgen Schmidhuber}.} \bibinfo{year}{1997}\natexlab{}.
\newblock \showarticletitle{Long short-term memory}.
\newblock \bibinfo{journal}{\emph{Neural computation}} \bibinfo{volume}{9},
  \bibinfo{number}{8} (\bibinfo{year}{1997}), \bibinfo{pages}{1735--1780}.
\newblock


\bibitem[\protect\citeauthoryear{Huang, Xu, and Yu}{Huang
  et~al\mbox{.}}{2015}]%
        {huang2015bidirectional}
\bibfield{author}{\bibinfo{person}{Zhiheng Huang}, \bibinfo{person}{Wei Xu},
  {and} \bibinfo{person}{Kai Yu}.} \bibinfo{year}{2015}\natexlab{}.
\newblock \showarticletitle{Bidirectional LSTM-CRF models for sequence
  tagging}.
\newblock \bibinfo{journal}{\emph{arXiv preprint arXiv:1508.01991}}
  (\bibinfo{year}{2015}).
\newblock


\bibitem[\protect\citeauthoryear{Kiefer and Wolfowitz}{Kiefer and
  Wolfowitz}{1956}]%
        {kiefer1956consistency}
\bibfield{author}{\bibinfo{person}{Jack Kiefer} {and} \bibinfo{person}{Jacob
  Wolfowitz}.} \bibinfo{year}{1956}\natexlab{}.
\newblock \showarticletitle{Consistency of the maximum likelihood estimator in
  the presence of infinitely many incidental parameters}.
\newblock \bibinfo{journal}{\emph{The Annals of Mathematical Statistics}}
  (\bibinfo{year}{1956}), \bibinfo{pages}{887--906}.
\newblock


\bibitem[\protect\citeauthoryear{Kingma and Ba}{Kingma and Ba}{2014}]%
        {kingma2014adam}
\bibfield{author}{\bibinfo{person}{Diederik~P Kingma} {and}
  \bibinfo{person}{Jimmy Ba}.} \bibinfo{year}{2014}\natexlab{}.
\newblock \showarticletitle{Adam: A method for stochastic optimization}.
\newblock \bibinfo{journal}{\emph{arXiv preprint arXiv:1412.6980}}
  (\bibinfo{year}{2014}).
\newblock


\bibitem[\protect\citeauthoryear{Kotsiantis}{Kotsiantis}{2007}]%
        {kotsiantis2007supervised}
\bibfield{author}{\bibinfo{person}{Sotiris~B Kotsiantis}.}
  \bibinfo{year}{2007}\natexlab{}.
\newblock \showarticletitle{Supervised machine learning: A review of
  classification techniques}.
\newblock  (\bibinfo{year}{2007}).
\newblock


\bibitem[\protect\citeauthoryear{Krizhevsky and Hinton}{Krizhevsky and
  Hinton}{2009}]%
        {krizhevsky2009learning}
\bibfield{author}{\bibinfo{person}{Alex Krizhevsky} {and}
  \bibinfo{person}{Geoffrey Hinton}.} \bibinfo{year}{2009}\natexlab{}.
\newblock \bibinfo{booktitle}{\emph{Learning multiple layers of features from
  tiny images}}.
\newblock \bibinfo{type}{{T}echnical {R}eport}.
  \bibinfo{institution}{Citeseer}.
\newblock


\bibitem[\protect\citeauthoryear{Li, Xu, Taylor, Studer, and Goldstein}{Li
  et~al\mbox{.}}{2018}]%
        {NIPS2018_7875}
\bibfield{author}{\bibinfo{person}{Hao Li}, \bibinfo{person}{Zheng Xu},
  \bibinfo{person}{Gavin Taylor}, \bibinfo{person}{Christoph Studer}, {and}
  \bibinfo{person}{Tom Goldstein}.} \bibinfo{year}{2018}\natexlab{}.
\newblock \showarticletitle{Visualizing the Loss Landscape of Neural Nets}.
\newblock In \bibinfo{booktitle}{\emph{Advances in Neural Information
  Processing Systems 31}}, \bibfield{editor}{\bibinfo{person}{S.~Bengio},
  \bibinfo{person}{H.~Wallach}, \bibinfo{person}{H.~Larochelle},
  \bibinfo{person}{K.~Grauman}, \bibinfo{person}{N.~Cesa-Bianchi}, {and}
  \bibinfo{person}{R.~Garnett}} (Eds.). \bibinfo{publisher}{Curran Associates,
  Inc.}, \bibinfo{pages}{6391--6401}.
\newblock


\bibitem[\protect\citeauthoryear{Li and Yuan}{Li and Yuan}{2017}]%
        {NIPS2017_6662}
\bibfield{author}{\bibinfo{person}{Yuanzhi Li} {and} \bibinfo{person}{Yang
  Yuan}.} \bibinfo{year}{2017}\natexlab{}.
\newblock \showarticletitle{Convergence Analysis of Two-layer Neural Networks
  with ReLU Activation}.
\newblock In \bibinfo{booktitle}{\emph{Advances in Neural Information
  Processing Systems 30}}, \bibfield{editor}{\bibinfo{person}{I.~Guyon},
  \bibinfo{person}{U.~V. Luxburg}, \bibinfo{person}{S.~Bengio},
  \bibinfo{person}{H.~Wallach}, \bibinfo{person}{R.~Fergus},
  \bibinfo{person}{S.~Vishwanathan}, {and} \bibinfo{person}{R.~Garnett}}
  (Eds.). \bibinfo{publisher}{Curran Associates, Inc.},
  \bibinfo{pages}{597--607}.
\newblock


\bibitem[\protect\citeauthoryear{Lin}{Lin}{2004}]%
        {lin2004note}
\bibfield{author}{\bibinfo{person}{Yi Lin}.} \bibinfo{year}{2004}\natexlab{}.
\newblock \showarticletitle{A note on margin-based loss functions in
  classification}.
\newblock \bibinfo{journal}{\emph{Statistics \& probability letters}}
  \bibinfo{volume}{68}, \bibinfo{number}{1} (\bibinfo{year}{2004}),
  \bibinfo{pages}{73--82}.
\newblock


\bibitem[\protect\citeauthoryear{Marcus, Marcinkiewicz, and Santorini}{Marcus
  et~al\mbox{.}}{1993}]%
        {marcus1993building}
\bibfield{author}{\bibinfo{person}{Mitchell~P Marcus},
  \bibinfo{person}{Mary~Ann Marcinkiewicz}, {and} \bibinfo{person}{Beatrice
  Santorini}.} \bibinfo{year}{1993}\natexlab{}.
\newblock \showarticletitle{Building a large annotated corpus of English: The
  Penn Treebank}.
\newblock \bibinfo{journal}{\emph{Computational linguistics}}
  \bibinfo{volume}{19}, \bibinfo{number}{2} (\bibinfo{year}{1993}),
  \bibinfo{pages}{313--330}.
\newblock


\bibitem[\protect\citeauthoryear{Millar}{Millar}{2011}]%
        {millar2011maximum}
\bibfield{author}{\bibinfo{person}{Russell~B Millar}.}
  \bibinfo{year}{2011}\natexlab{}.
\newblock \bibinfo{booktitle}{\emph{Maximum likelihood estimation and
  inference: with examples in R, SAS and ADMB}}. Vol.~\bibinfo{volume}{111}.
\newblock \bibinfo{publisher}{John Wiley \& Sons}.
\newblock


\bibitem[\protect\citeauthoryear{Nair and Hinton}{Nair and Hinton}{2010}]%
        {nair2010rectified}
\bibfield{author}{\bibinfo{person}{Vinod Nair} {and}
  \bibinfo{person}{Geoffrey~E Hinton}.} \bibinfo{year}{2010}\natexlab{}.
\newblock \showarticletitle{Rectified linear units improve restricted boltzmann
  machines}. In \bibinfo{booktitle}{\emph{Proceedings of the 27th international
  conference on machine learning (ICML-10)}}. \bibinfo{pages}{807--814}.
\newblock


\bibitem[\protect\citeauthoryear{Pennington, Socher, and Manning}{Pennington
  et~al\mbox{.}}{2014}]%
        {pennington2014glove}
\bibfield{author}{\bibinfo{person}{Jeffrey Pennington},
  \bibinfo{person}{Richard Socher}, {and} \bibinfo{person}{Christopher
  Manning}.} \bibinfo{year}{2014}\natexlab{}.
\newblock \showarticletitle{Glove: Global vectors for word representation}. In
  \bibinfo{booktitle}{\emph{Proceedings of the 2014 conference on empirical
  methods in natural language processing (EMNLP)}}.
  \bibinfo{pages}{1532--1543}.
\newblock


\bibitem[\protect\citeauthoryear{Pires and Szepesv{\'a}ri}{Pires and
  Szepesv{\'a}ri}{2016}]%
        {pires2016multiclass}
\bibfield{author}{\bibinfo{person}{Bernardo~{\'A}vila Pires} {and}
  \bibinfo{person}{Csaba Szepesv{\'a}ri}.} \bibinfo{year}{2016}\natexlab{}.
\newblock \showarticletitle{Multiclass classification calibration functions}.
\newblock \bibinfo{journal}{\emph{arXiv preprint arXiv:1609.06385}}
  (\bibinfo{year}{2016}).
\newblock


\bibitem[\protect\citeauthoryear{Schuster and Paliwal}{Schuster and
  Paliwal}{1997}]%
        {schuster1997bidirectional}
\bibfield{author}{\bibinfo{person}{Mike Schuster} {and}
  \bibinfo{person}{Kuldip~K Paliwal}.} \bibinfo{year}{1997}\natexlab{}.
\newblock \showarticletitle{Bidirectional recurrent neural networks}.
\newblock \bibinfo{journal}{\emph{IEEE Transactions on Signal Processing}}
  \bibinfo{volume}{45}, \bibinfo{number}{11} (\bibinfo{year}{1997}),
  \bibinfo{pages}{2673--2681}.
\newblock


\bibitem[\protect\citeauthoryear{Smith, Everhart, Dickson, Knowler, and
  Johannes}{Smith et~al\mbox{.}}{1988}]%
        {smith1988using}
\bibfield{author}{\bibinfo{person}{Jack~W Smith}, \bibinfo{person}{JE
  Everhart}, \bibinfo{person}{WC Dickson}, \bibinfo{person}{WC Knowler}, {and}
  \bibinfo{person}{RS Johannes}.} \bibinfo{year}{1988}\natexlab{}.
\newblock \showarticletitle{Using the ADAP learning algorithm to forecast the
  onset of diabetes mellitus}. In \bibinfo{booktitle}{\emph{Proceedings of the
  Annual Symposium on Computer Application in Medical Care}}. American Medical
  Informatics Association, \bibinfo{pages}{261}.
\newblock


\bibitem[\protect\citeauthoryear{Socher, Perelygin, Wu, Chuang, Manning, Ng,
  and Potts}{Socher et~al\mbox{.}}{2013}]%
        {socher2013recursive}
\bibfield{author}{\bibinfo{person}{Richard Socher}, \bibinfo{person}{Alex
  Perelygin}, \bibinfo{person}{Jean Wu}, \bibinfo{person}{Jason Chuang},
  \bibinfo{person}{Christopher~D Manning}, \bibinfo{person}{Andrew Ng}, {and}
  \bibinfo{person}{Christopher Potts}.} \bibinfo{year}{2013}\natexlab{}.
\newblock \showarticletitle{Recursive deep models for semantic compositionality
  over a sentiment treebank}. In \bibinfo{booktitle}{\emph{Proceedings of the
  2013 conference on empirical methods in natural language processing}}.
  \bibinfo{pages}{1631--1642}.
\newblock


\bibitem[\protect\citeauthoryear{Srivastava, Hinton, Krizhevsky, Sutskever, and
  Salakhutdinov}{Srivastava et~al\mbox{.}}{2014}]%
        {JMLR:v15:srivastava14a}
\bibfield{author}{\bibinfo{person}{Nitish Srivastava},
  \bibinfo{person}{Geoffrey Hinton}, \bibinfo{person}{Alex Krizhevsky},
  \bibinfo{person}{Ilya Sutskever}, {and} \bibinfo{person}{Ruslan
  Salakhutdinov}.} \bibinfo{year}{2014}\natexlab{}.
\newblock \showarticletitle{Dropout: A Simple Way to Prevent Neural Networks
  from Overfitting}.
\newblock \bibinfo{journal}{\emph{Journal of Machine Learning Research}}
  \bibinfo{volume}{15} (\bibinfo{year}{2014}), \bibinfo{pages}{1929--1958}.
\newblock
\urldef\tempurl%
\url{http://jmlr.org/papers/v15/srivastava14a.html}
\showURL{%
\tempurl}


\bibitem[\protect\citeauthoryear{Tai, Socher, and Manning}{Tai
  et~al\mbox{.}}{2015}]%
        {tai2015improved}
\bibfield{author}{\bibinfo{person}{Kai~Sheng Tai}, \bibinfo{person}{Richard
  Socher}, {and} \bibinfo{person}{Christopher~D Manning}.}
  \bibinfo{year}{2015}\natexlab{}.
\newblock \showarticletitle{Improved semantic representations from
  tree-structured long short-term memory networks}.
\newblock \bibinfo{journal}{\emph{arXiv preprint arXiv:1503.00075}}
  (\bibinfo{year}{2015}).
\newblock


\bibitem[\protect\citeauthoryear{Tjong Kim~Sang and De~Meulder}{Tjong Kim~Sang
  and De~Meulder}{2003}]%
        {tjong2003introduction}
\bibfield{author}{\bibinfo{person}{Erik~F Tjong Kim~Sang} {and}
  \bibinfo{person}{Fien De~Meulder}.} \bibinfo{year}{2003}\natexlab{}.
\newblock \showarticletitle{Introduction to the CoNLL-2003 shared task:
  Language-independent named entity recognition}. In
  \bibinfo{booktitle}{\emph{Proceedings of the seventh conference on Natural
  language learning at HLT-NAACL 2003-Volume 4}}. Association for Computational
  Linguistics, \bibinfo{pages}{142--147}.
\newblock


\bibitem[\protect\citeauthoryear{Vapnik}{Vapnik}{2013}]%
        {vapnik2013nature}
\bibfield{author}{\bibinfo{person}{Vladimir Vapnik}.}
  \bibinfo{year}{2013}\natexlab{}.
\newblock \bibinfo{booktitle}{\emph{The nature of statistical learning
  theory}}.
\newblock \bibinfo{publisher}{Springer science \& business media}.
\newblock


\bibitem[\protect\citeauthoryear{Wahba et~al\mbox{.}}{Wahba
  et~al\mbox{.}}{1999}]%
        {wahba1999support}
\bibfield{author}{\bibinfo{person}{Grace Wahba} {et~al\mbox{.}}}
  \bibinfo{year}{1999}\natexlab{}.
\newblock \showarticletitle{Support vector machines, reproducing kernel Hilbert
  spaces and the randomized GACV}.
\newblock \bibinfo{journal}{\emph{Advances in Kernel Methods-Support Vector
  Learning}}  \bibinfo{volume}{6} (\bibinfo{year}{1999}),
  \bibinfo{pages}{69--87}.
\newblock


\bibitem[\protect\citeauthoryear{Weiss and Kapouleas}{Weiss and
  Kapouleas}{1990}]%
        {weiss1990empirical}
\bibfield{author}{\bibinfo{person}{Sholom~M Weiss} {and}
  \bibinfo{person}{Ioannis Kapouleas}.} \bibinfo{year}{1990}\natexlab{}.
\newblock \showarticletitle{An empirical comparison of pattern recognition,
  neural nets and machine learning classification methods}.
\newblock \bibinfo{journal}{\emph{Readings in machine learning}}
  (\bibinfo{year}{1990}), \bibinfo{pages}{177--183}.
\newblock


\bibitem[\protect\citeauthoryear{Witten, Frank, Hall, and Pal}{Witten
  et~al\mbox{.}}{2016}]%
        {witten2016data}
\bibfield{author}{\bibinfo{person}{Ian~H Witten}, \bibinfo{person}{Eibe Frank},
  \bibinfo{person}{Mark~A Hall}, {and} \bibinfo{person}{Christopher~J Pal}.}
  \bibinfo{year}{2016}\natexlab{}.
\newblock \bibinfo{booktitle}{\emph{Data Mining: Practical machine learning
  tools and techniques}}.
\newblock \bibinfo{publisher}{Morgan Kaufmann}.
\newblock


\bibitem[\protect\citeauthoryear{Wu and Liu}{Wu and Liu}{2007}]%
        {wu2007robust}
\bibfield{author}{\bibinfo{person}{Yichao Wu} {and} \bibinfo{person}{Yufeng
  Liu}.} \bibinfo{year}{2007}\natexlab{}.
\newblock \showarticletitle{Robust truncated hinge loss support vector
  machines}.
\newblock \bibinfo{journal}{\emph{J. Amer. Statist. Assoc.}}
  \bibinfo{volume}{102}, \bibinfo{number}{479} (\bibinfo{year}{2007}),
  \bibinfo{pages}{974--983}.
\newblock


\end{thebibliography}

\end{document}